\pdfoutput=1

\documentclass[11pt]{article}

\usepackage[]{emnlp2021}

\usepackage{times}
\usepackage{latexsym}

\usepackage[T1]{fontenc}

\usepackage[utf8]{inputenc}

\usepackage{microtype}

\usepackage{amsmath}
\usepackage{graphicx}
\usepackage{makecell} 

\title{\emph{Different Strokes for Different Folks}:  Investigating Appropriate Further Pre-training Approaches for Diverse Dialogue Tasks}

\author{Yao Qiu, \ Jinchao Zhang \thanks{\ \ Jinchao Zhang is the corresponding author.}, \ Jie Zhou \\
  Pattern Recognition Center, WeChat AI, Tencent Inc, China \\
  \texttt{\{yasinqiu, dayerzhang, withtomzhou\}@tencent.com}
}

\begin{document}
\maketitle

\begin{abstract}
Loading models pre-trained on the large-scale corpus in the general domain and fine-tuning them on specific downstream tasks is gradually becoming a paradigm in Natural Language Processing.  
Previous investigations prove that introducing a \textbf{further pre-training} phase between pre-training and fine-tuning phases to adapt the model on the domain-specific unlabeled data can bring positive effects.
However, most of these further pre-training works just keep running the conventional pre-training task, e.g., masked language model, which can be regarded as the domain adaptation to bridge the data distribution gap.
After observing diverse downstream tasks, we suggest that different tasks may also need a further pre-training phase with appropriate training tasks to bridge the task formulation gap. 
To investigate this, we carry out a study for improving multiple task-oriented dialogue downstream tasks through designing various tasks at the further pre-training phase.
The experiment shows that different downstream tasks prefer different further pre-training tasks, which have intrinsic correlation and most further pre-training tasks significantly improve certain target tasks rather than all. 
Our investigation indicates that it is of great importance and effectiveness to design appropriate further pre-training tasks modeling specific information that benefit downstream tasks.
Besides, we present multiple constructive empirical conclusions for enhancing task-oriented dialogues.
\end{abstract}

\section{Introduction}
Pre-trained models, e.g., BERT~\cite{devlin2019bert}, RoBERTa~\cite{liu2019roberta} and GPT2~\cite{radford2019language}, have been widely used in many NLP tasks. These models are pre-trained on the large-scale general text corpus, such as Wikipedia or books, with self-supervised training objectives. Fine-tuning these models on downstream tasks can achieve excellent performance. 

Recently, \newcite{gururangan2020don} proposed a domain-adaptive pre-training method, they further pre-training the RoBERTa on a large corpus of unlabeled domain-specific text, e.g., biomedical papers and computer science papers, before fine-tuning on downstream tasks and achieved strong performance. Besides, they proved that it is also helpful to continue pre-training on the task-specific text. 
\newcite{wu2020tod} applied this method to task-oriented dialogue and proposed a new self-supervised pre-training objective on dialogue corpus. 
Despite they achieved performance improvements, the improvements on different downstream tasks vary a lot, some tasks even obtain no improvement, which indicates that different downstream tasks may need different further pre-training tasks. 

To investigate this issue, we carry out experiments in the area of task-oriented dialogue. 
We choose one popular pre-training language model, BERT~\cite{devlin2019bert} as our base model, 
and construct a large scale domain-specific dialogue corpus which consists of nine task-oriented datasets for further pre-training~\cite{wu2020tod}.
We also select four core task-oriented dialogue tasks, intent recognition, dialogue action prediction, response selection, and dialog state tracking as the downstream tasks used in fine-tuning phase.
We aim to explore the following questions: 
1) In the area of task-oriented dialogue, can further pre-training using the masked language model improve the performance of all downstream tasks? 
2) Do different further pre-training tasks have different effects on different downstream tasks? 
3) Which factors affect whether a further pre-training task can achieve improvement on a certain downstream task?
4) Does combining different further pre-training tasks benefits more downstream tasks?

To answer these questions, we design five self-supervised pre-training tasks according to different characteristics of the downstream tasks. 
Specifically, we first use specially designed pre-training tasks to further pre-training BERT on the domain-specific corpus, obtaining multiple new pre-trained models, denoted as BERT's variants. 
Then, we fine-tune these variants on all downstream tasks and observe the effect of different pre-training tasks on different downstream tasks.
From experiment results, we figure out that:
1) Further pre-training with masked language model does not achieve improvements for all downstream tasks, it is necessary to design special further pre-training tasks according to the characteristics of dialogue data.
2) Different pre-training tasks do have different effects on different downstream tasks, and there is a need to design a specific pre-training task for a certain downstream task.
3) Model's ability and structure are two key factors influencing effectiveness of the further pre-training on a certain downstream task.
4) Training two further pre-training tasks in a multi-task paradigm does not lead to incremental performance improvements on downstream tasks.

The main contribution of our work is to give a set of empirical principles about how to design effective further pre-training tasks for enhancing the task-oriented dialogue. 
The key points of the design are to make the model structures of the pre-training task and the downstream task similar and let the model learn the abilities required by downstream tasks in the pre-training phase while maintaining the masked language model's training. We release the source code at the GitHub repo. \footnote{https://github.com/FFYYang/DSDF.git}

\section{Background}

\subsection{Pre-trained Models}
\paragraph{Two-stage Training.}
Large pre-training models, such as BERT~\cite{devlin2018bert}, RoBERTa~\cite{liu2019roberta}, GPT2~\cite{radford2019language}, XLNet~\cite{yang2019xlnet} T5~\cite{raffel2019exploring}, are trained on massive general domain text with self-supervised training objectives, like masked language model~\cite{devlin2018bert} and permutation language model~\cite{yang2019xlnet}. These models learned strong and general word representations, fine-tuning these pre-trained models on downstream tasks is proved to be effective.

\paragraph{Three-stage Training.}
Recently, further pre-training large language models on domain-specific corpus before fine-tuning on downstream tasks has become a popular and effective paradigm. 
\newcite{gururangan2020don} proposed domain-adaptive pre-training and task-adaptive pre-training methods, and they proved that such a second phase of pre-training in a specific domain leads to performance gains. 
\newcite{wu2020tod} applied the second phase pre-training on task-oriented dialogue, in addition to masked language modeling objective, they also proposed a new self-supervised objective according to the characteristic of dialogue corpus. 
However, the performance improvement gained from their proposed methods varies a lot across different downstream tasks, which indicates different downstream tasks may need different further pre-training tasks rather than the conventional one, such as MLM.

\subsection{Task-oriented Dialogue}
A task-oriented dialog system aims to assist the user in completing certain tasks in one or several specific domains, such as restaurant booking, weather query, and flight booking.
The entire system usually consists of four modules, including natural language understanding (NLU), dialog state tracking (DST), dialog policy, and natural language generation (NLG).
In this work, we focus on four core tasks:
\begin{itemize}
    \item \textbf{Intent recognition}: The model is required to predict the intent type given the user utterance. Intent type is a high-level classification label of the user utterance, such as \textit{Query} and \textit{Inform}, which indicates the function of the user utterance.
    \item \textbf{Dialog act prediction}: The model is required to predict the dialog act (e.g., \textit{Question}, \textit{Statement}) of the next response given the whole dialog history.
    \item \textbf{Response selection}: The model is required to select the proper response from many candidate responses given the whole dialog history. The negative candidate responses are randomly sampled.
    \item \textbf{Dialog state tracking}: The dialog state tracker estimates the user’s goal in each time step by taking the entire dialog context as input. The dialog state at time $t$ can be regarded as an abstracted representation of the previous turns until $t$.
\end{itemize}

\begin{figure*}[]
\centering
\includegraphics[width=16cm]{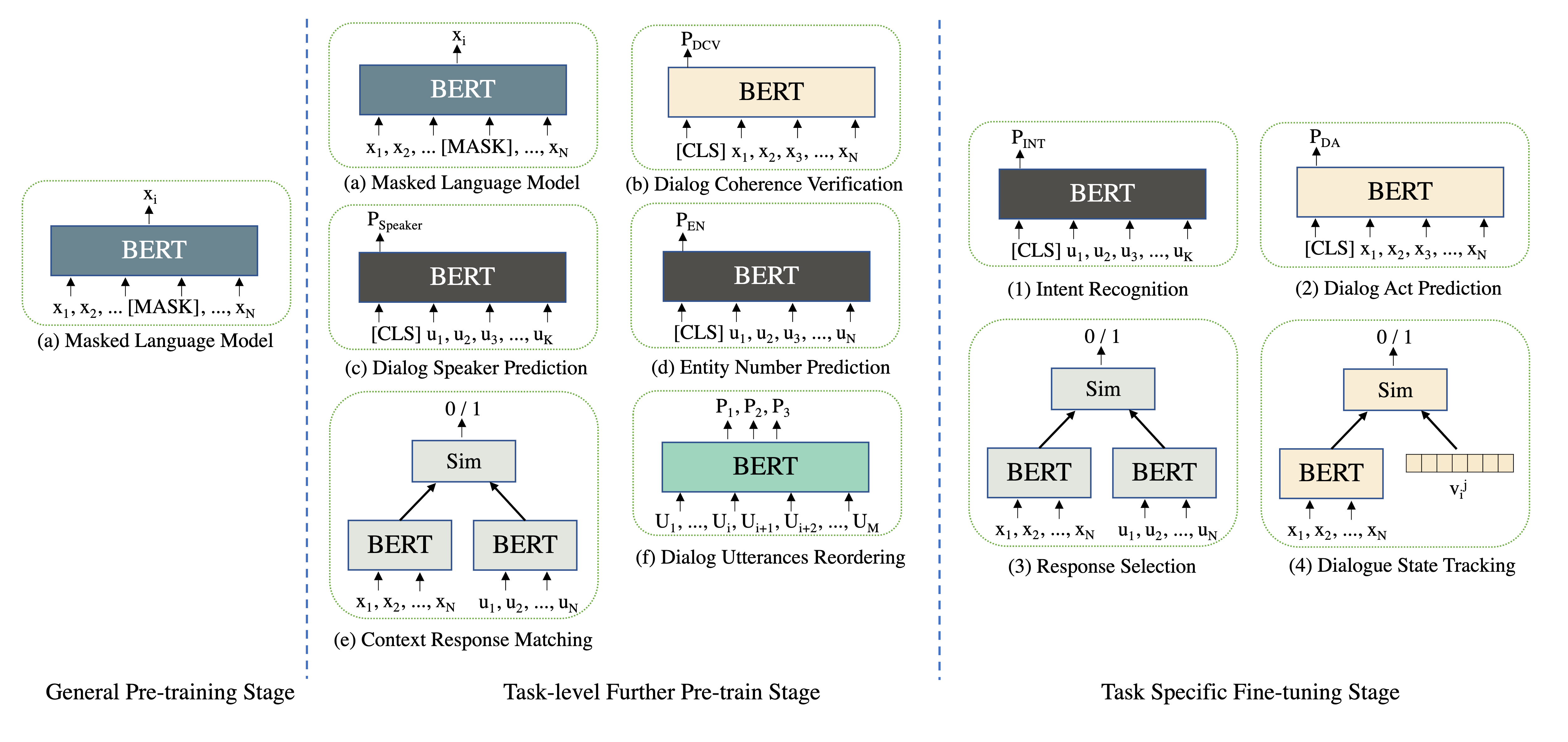}
\caption{The pipeline of our training process, started from a pre-trained BERT, then further pre-train BERT on dialogue corpus using multiple proposed tasks, the resulting models are fine-tuned on downstream tasks.}
\label{exp_pipeline_2}
\end{figure*}

\section{Approaches}
In this section, we firstly present the three-stage training framework,
then introduce five expressly designed further pre-training tasks and downstream tasks.
At last, we present some heuristic analysis on the relations between the tasks in the further pre-training and the fine-tuning stage.

\subsection{Three-stage Training for the Task-oriented Dialogue}
We design a three-stage training framework, includes the \textbf{general pre-training stage}, \textbf{task-level further pre-training stage} and the \textbf{task-specific fine-tuning stage} for enhancing the various tasks in the task-oriented dialogue, as shown in Figure~\ref{exp_pipeline_2}.
The general pre-training stage aims to learn general word representation. 
The task-level further pre-training stage contains multiple optional tasks trained on the un-labeled dialogue corpus. 
The task-specific fine-tuning stage is to train specific models for solving the downstream task such as intent recognition. 
To be emphasized, our further pre-training stage attempts to bridge the task-level gap between the pre-training and the fine-tuning stage rather than the domain adaptation on the data-level~\cite{gururangan2020don}.

\subsection{Task-level Further Pre-training}
To enhance the task-oriented dialogue through bridging the task-level gap between pre-training and fine-tuning, 
we design multiple optional tasks which can be trained on dialogue corpus without any human annotation.

\paragraph{Dialog Speaker Prediction (DSP).}
The model is required to predict the speaker (user or agent) of a given utterance. The model can learn a better single utterance representation from this task. The input of the model is a single utterance $U = u_1, u_2, ..., u_K$, where $K$ is the utterance length. The model outputs a binary result indicating the speaker is a user or agent.
\begin{equation}
    P_{Speaker} = Softmax(W_{DSP} \cdot F_{bert}(U) )
\end{equation}
Where $F_{bert}$ is the forward function of BERT, we use its [CLS] representations as the utterance representation. $W_{DSP}$ is a trainable linear mapping matrix. The task is trained with the cross-entropy loss.

\paragraph{Context Response Matching (CRM).}
Given a dialog context, the model selects the proper response from many randomly sampled candidate responses. 
This task is in the same as the response contrastive loss proposed by \newcite{wu2020tod}.
The model can learn the dialogue coherence information from this task.

\paragraph{Dialogue Coherence Verification (DCV).}
This task asks the model to predict whether a dialog is coherent. The incoherent dialog is constructed by randomly replacing some utterances in the dialog. The model can learn a better multi-turn dialog representation from this task.
We first randomly select half of the training data and randomly replace some utterances in the dialogue to destroy their coherence. 
The input of the model is the whole dialog, concatenating all utterances together, denoted as $S=x_1, x_2, ..., x_N$, where $N$ is the sequence length. The model outputs a binary prediction.
\begin{equation}
    P_{DCV} = Softmax(W_{DCV} \cdot F_{bert}(S))
\end{equation}
Where $F_{bert}$ is the forward function of BERT, we use its [CLS] representations as the dialog representation. $W_{DCV}$ is a trainable linear mapping matrix. The task is trained with the cross-entropy loss.

\paragraph{Entity Number Prediction (ENP).} 
The model predicts the number of entities contained in an utterance. Entities are extracted using the open-source tool \textit{stanza}~\footnote{https://github.com/stanfordnlp/stanza}. The model can learn a better single utterance representation and entity information.
This task is formulated as a multi-class classification problem, where we input a single utterance $U$, and the model predicts one single class indicating how many entities are contained in the utterance.
\begin{equation}
    P_{EN} = Softmax(W_{ENP} \cdot F_{bert}(U))
\end{equation}
Where $F_{bert}$ is the forward function of BERT. $W_{ENP}$ is a trainable linear mapping matrix. The task is trained with the cross-entropy loss.

\paragraph{Dialog Utterances Reordering (DUR).}
The model reorders a group of shuffled utterances. The model can learn dialog coherence information from this task.
The input of the model is the whole dialog, but some utterances' positions are shuffled.
We put special tokens [USR] and [SYS] at the front of each utterance indicating it is spoken by a user or agent. 
We concatenate all utterances together, feed them to BERT, and take the representation of [USR] and [SYS] as the representation of each utterance. 
The model predicts the correct relative position of the shuffled utterances. 
For example, utterances $U_i, U_{i+1}, U_{i+2}$ are shuffled, we first use BERT to get their representations $R_{i}, R_{i+1}, R_{i+2}$, and use a FFN and softmax function to get the probability distribution of their relative positions, $y_p = [y_p^1, y_p^2, y_p^3]$. The loss is calculated as: 
\begin{equation}
L_{DUR} = Avg(-SUM(y_t * log(y_p+eps)))
\end{equation}
Where $y_p$ is the correct probability distribution of these utterances relative positions, for example, suppose the correct relative position is $[2,1,3]$, then $y_p = Softmax([2,1,3])$.

\begin{table*}[]
\small
\centering
\begin{tabular}{l| cccc| cccc}
\hline
 & \multicolumn{4}{c| }{\textbf{Abilities}} & \multicolumn{4}{c }{\textbf{Structures}} \\
 & \makecell[c]{Single Turn \\ Representation}  & \makecell[c]{Multi Turn \\ Representation} &\makecell[c]{Coherence}  & \makecell[c]{Entity\\Information} & \makecell[c]{Single Turn \\ Classifier} & \makecell[c]{Multi Turn \\ Classifier} & \makecell[c]{Siamese \\ Model} & \makecell[c]{Rank \\ Loss} \\
\hline
INT & ※ &  &  & ※ & ※ &  &  &  \\
DA &  & ※ &  &  &  & ※ &  &  \\
RS &  &  & ※ &  &  &  & ※ &  \\
DST &  & ※ &  & ※ &  & ※ &  &  \\
\hline
\hline
DSP & ※ &  &  &  & ※ &  &  &  \\
CRM &  &  & ※ &  &  &  & ※ &  \\
DCV &  & ※ &  &  &  & ※ &  &  \\
ENP & ※ &  &  & ※ & ※ &  &  &  \\
DUR &  &  & ※ &  &  &  &  & ※ \\
\hline
\end{tabular}
\caption{This table shows the comparison between further pre-training and downstream tasks from the ability and structure perspective. The above four tasks are downstream tasks, the below five tasks are further pre-training tasks. ※ indicates the task has the ability or belongs to the model structure.}
\label{relation}
\end{table*}

\subsection{Task Specific Fine-tuning}
After further pre-training, we fine-tuning our models on each downstream task individually. These downstream tasks are modeled in different forms following~\cite{wu2020tod}.

\paragraph{Intent Recognition (INT).}
The task is a multi-class classification problem, 
the input of the model is a single utterance $U$ and model predicts one single intent type.
\begin{equation}
    P_{INT} = Softmax(W_{INT} \cdot F_{bert}(U)) ,
\end{equation}
The task is trained with the cross-entropy loss.

\paragraph{Dialogue Act Prediction (DA).} 
The task is modeled as a multi-label classification problem, since a system response may contain multiple dialogue acts. 
The model's input is the whole dialogue history $S$, and the model outputs a binary prediction for each possible dialogue act.
\begin{equation}
    P_{DA} = Sigmoid(W_{DA} \cdot F_{bert}(S)), 
\end{equation}
It is trained with the binary cross-entropy loss.

\paragraph{Response Selection (RS).}
The model selects the most proper system response from multiple candidates. We utilize a siamese structure and compute similarity scores between dialogue history $H$ and a candidate response $R_i$.
\begin{equation}
    s_i = Sim(F_{bert}(H), F_{bert}(R_i)), 
\end{equation}
Where $s_i$ is the cosine similarity.
The negative candidates are randomly sampled from the corpus.

\paragraph{Dialogue State Tracking (DST)} 
is modeled as a multi-class classification task based on a predefined ontology. 
The model's input is the whole dialogue history $S$, and the model predicts the value of the slot for each (domain, slot) pair. We define $v_i^j$ as the $i$-th value for $j$-th (domain, slot) pair, we use BERT to obtain its representation which is fixed during the whole fine-tuning stage.
\begin{equation}
    S_i^j = Sim(G_j(F_{bert}(X)), F_{bert}(v_i^j)),
\end{equation}
Where $Sim$ is the cosine similarity function, and $S^j$ is the probability distribution of the $j$-th (domain, slot) pair over its possible values. $G_j$ is the slot projection layer of the $j$-th (domain, slot) pair, and the number of layers $|G|$ is equal to the number of (domain, slot) pairs. The task is trained with the cross-entropy loss summed over all the pairs.

All of the proposed tasks are trained with the masked language model in a multi-task paradigm. In addition, these tasks are optional,
we focus on investigating their relations with each downstream task.

\subsection{Heuristic Analysis on Task Relations between Further Pre-training and Fine-tuning}

We analyse the task relations from two perspectives: model \textit{ability} and \textit{structure}.
Ability refers to the information or knowledge the model learns, for example, the ability of single turn representation, the knowledge about the entity.
Structure refers to the model's network structure and its objective function, for example, the siamese structure and list-wise ranking loss function.
The details are shown in Table~\ref{relation}. We suggest that if a further pre-training task learns similar abilities or has a similar model structure the with the downstream task, then the further pre-training will be more effective for fine-tuning.

\begin{table*}[]
\small
\begin{tabular}{l| llllll| llll}
\hline 
 & \multicolumn{6}{c|}{\textbf{DA}} & \multicolumn{4}{c}{\textbf{INT}} \\
 & \multicolumn{2}{c}{MWOZ} & \multicolumn{2}{c}{DSTC2} & \multicolumn{2}{c|}{GSIM} & Acc & Acc & Acc & Recall \\
 & $f1_{micro}$ & $f1_{macro}$ & $f1_{micro}$ & $f1_{macro}$ & $f1_{micro}$ & $f1_{macro}$ & (all) & (in) & (out) & (out)  \\
\hline
$BERT^2$ & 90.90 & 81.31 & 91.16 & 37.67 & 99.07 & 45.49 & 84.96 & 95.20 & 88.70 & 38.87 \\
$MLM^3$ & \textbf{91.51} & 79.77 & 87.76 & 36.99 & \textbf{99.35} & \textbf{45.70} & \textbf{85.10} & \textbf{95.90} & 88.30 & 36.70 \\
\hline
\hline
 & \multicolumn{6}{c|}{\textbf{RS}} & \multicolumn{2}{c}{\textbf{DST}} &  &  \\
 & \multicolumn{2}{c}{MWOZ} & \multicolumn{2}{c}{DSTC2} & \multicolumn{2}{c|}{GSIM} & $acc_{joint}$ & $acc_{slot}$ &  &  \\
 & $R_{100}@1$ & $R_{100}@3$ & $R_{100}@1$ & $R_{100}@3$ & $R_{100}@1$ & $R_{100}@3$ &  &  &  &  \\
\hline
$BERT^2$ & 47.39 & 74.46 & 48.33 & 63.30 & 19.21 & 40.02 & 47.96 & 96.86 &  &  \\
$MLM^3$ & \textbf{59.55} & \textbf{82.63} & \textbf{54.25} & \textbf{68.91} & \textbf{35.71} & \textbf{58.73} & \textbf{48.72} & \textbf{96.94} &  & \\
\hline
\end{tabular}
\caption{The results of the experiment investigating the effect of data-level further pre-train. $BERT^2$ does not contain a further pre-training stage, $MLM^3$ utilizes masked language model to further pre-train BERT on un-labeled dialogue corpus before fine-tuning. $MLM^3$ does not surpass $BERT^2$ in all metrics and datasets.}
\label{bert_mlm}
\end{table*}

\begin{table*}[]
\small
\begin{tabular}{l| llllll| llll}
\hline 
 & \multicolumn{6}{c|}{\textbf{DA}} & \multicolumn{4}{c}{\textbf{INT}} \\
 & \multicolumn{2}{c}{MWOZ} & \multicolumn{2}{c}{DSTC2} & \multicolumn{2}{c|}{GSIM} & Acc & Acc & Acc & Recall \\
 & $f1_{micro}$ & $f1_{macro}$ & $f1_{micro}$ & $f1_{macro}$ & $f1_{micro}$ & $f1_{macro}$ & (all) & (in) & (out) & (out)  \\
\hline
$MLM^3$ & 91.51 & 79.77 & 87.76 & 36.99 & 99.35 & 45.70 & 85.10 & \textbf{95.90} & 88.30 & 36.70 \\
DSP & 91.46 & \textbf{80.87} & 93.26 & \textbf{43.54} & 99.35 & 45.72 & \textbf{85.78} & 95.52 & \textbf{89.28} & \textbf{41.97} \\
CRM & 91.52 & 79.87 & 92.59 & 40.48 & 99.31 & 45.68 & 85.02 & 94.91 & 88.92 & 40.53 \\
DCV & \textbf{91.67} & 80.47 & \textbf{93.84} & 42.44 & \textbf{99.41} & \textbf{45.77} & 85.00 & 95.76 & 88.32 & 36.60 \\
\hline
\hline
 & \multicolumn{6}{c|}{\textbf{RS}} & \multicolumn{2}{c}{\textbf{DST}} &  &  \\
 & \multicolumn{2}{c}{MWOZ} & \multicolumn{2}{c}{DSTC2} & \multicolumn{2}{c|}{GSIM} & $acc_{joint}$ & $acc_{slot}$ &  &  \\
 & $R_{100}@1$ & $R_{100}@3$ & $R_{100}@1$ & $R_{100}@3$ & $R_{100}@1$ & $R_{100}@3$ &  &  &  &  \\
\hline
$MLM^3$ & 59.55 & 82.63 & 54.25 & 68.91 & 35.71 & 58.73 & 48.72 & 96.94 &  &  \\
DSP & 61.79 & 84.04 & 53.06 & 66.94 & 36.59 & 59.37 & 49.02 & 96.96 &  &  \\
CRM & \textbf{64.27} & \textbf{86.15} & \textbf{58.12} & \textbf{71.86} & \textbf{41.94} & \textbf{66.47} & 48.89 & 96.97 &  &  \\
DCV & 60.14 & 83.51 & 54.57 & 68.45 & 29.35 & 53.12 & \textbf{51.18} & \textbf{97.15} &  & \\
\hline
\end{tabular}
\caption{Results of the experiment investigating the effect of further pre-training tasks. These three tasks outperform MLM on most metrics, and different further pre-training tasks benefit to different downstream tasks.}
\label{question2}
\end{table*}

\section{Experimental Setup}

\subsection{Dialogue Datasets for Further Pre-training}
Following \newcite{wu2020tod}, we construct the further pre-training dataset by combining nine different multi-turn goal-oriented datasets (Frames~\cite{asri2017frames}, MetaLWOZ~\cite{lee2019multi}, WOZ~\cite{mrkvsic2016neural}, CamRest676~\cite{wen2016network}, MSR-E2E~\cite{li2018microsoft}, MWOZ~\cite{budzianowski2018multiwoz}, Schema~\cite{rastogi2020towards}, SMD~\cite{eric2017key} and Taskmaster~\cite{byrne2019taskmaster}). 
In total, there are 100,707 dialogues containing 1,388,152 utterances over 60 domains. 

\subsection{Evaluation Datasets}

We select four datasets, OOS, DSTC2, GSIM, and MWOZ, for downstream evaluation. Details of each evaluation dataset are discussed below.

\paragraph{OOS.}~\cite{larson2019evaluation} 
It contains 151 intent types across ten domains, including 150 in-scope and one out-of-scope intent.

\paragraph{DSTC2.}~\cite{henderson2014second} 
It is a machine-human task-oriented dataset, We follow \newcite{wu2020tod} to map the original dialogue act labels to universal dialogue acts, resulting in 19 acts.

\paragraph{MWOZ.}~\cite{budzianowski2018multiwoz} 
It is a popular benchmark for task-oriented dialogues. It has 30 (domain, slot) pairs across seven different domains. We use the revised version MWOZ 2.1.

\paragraph{GSIM.}~\cite{shah2018bootstrapping} 
It is a human-rewrote task-oriented dataset. Following \newcite{wu2020tod} we combine movie and restaurant domains into one single corpus, and map its dialogue act labels to universal dialogue acts, resulting in 13 acts.

\subsection{Training Setting} 


For further pre-training, we set the learning rate equal to 5e-5, batch size to 32, and maximum sequence length to 512.
For fine-tuning, we set the learning rate to 5e-5 (except dialog state tracking task, which is 3e-5). We use the batch size that maximizes the GPU usage.
We train our models using the Adam optimizer.
Models are early-stopped using the loss of a validation set. 
We train each downstream task three times with different seeds. We use 4 NVIDIA V100 GPUs for further pre-training and one for fine-tuning.
Our code is based on Transformers~\footnote{https://github.com/huggingface/transformers}

\section{Results and Discussion}
In this section, we collect experimental results and analyse the effects of different further pre-training tasks on different downstream tasks.

\subsection{Effect of the Data-level Further Pre-training}
To investigate the effect of the data-level further pre-training, we firstly further pre-train BERT with masked language model (MLM) on the un-labeled task-oriented dialogue corpus, then fine-tune the model on each downstream task, we denote this experiment as $MLM^3$. In contrast, we also directly fine-tune BERT on downstream tasks, the experiment is denoted as $BERT^2$.
The experiment results are shown in Table~\ref{bert_mlm}, $MLM^3$ outperforms $BERT^2$ on response selection and dialog state tracking task, as for dialog act prediction and intent recognition task, $MLM^3$ does not surpass $BERT^2$ in all metrics and datasets.
From the result, we can conclude that \textit{further pre-training using MLM objective does not bring performance improvement for all downstream tasks}, so it is necessary to design special further pre-training tasks according to the characteristics of the dialogue data.

\subsection{Effect of Various Further Pre-training Tasks}
To investigate the effects of different further pre-training tasks on different downstream tasks, we compare three further pre-training tasks, dialogue speaker prediction (DSP), context response matching (CRM), and dialogue coherence verification (DCV), each of which has special characteristics.
From the experiment results shown in Table~\ref{question2},
DSP, CRM, and DCV are better than $MLM^3$ on most of the metrics, this indicates the effectiveness of these auxiliary pre-training tasks.
In addition, we also observe that different pre-training tasks are more beneficial to different downstream tasks, for example, DSP is more beneficial to downstream intent recognition task than others, CRM is mainly beneficial to response selection, DCV is beneficial to dialogue act prediction and dialogue state tracking.
We can conclude that \textit{different pre-training tasks do have different effects on different downstream tasks}, so there is a need to design a specific pre-training task for a downstream task.

\begin{table*}[]
\small
\begin{tabular}{l| llllll| llll}
\hline 
 & \multicolumn{6}{c|}{\textbf{DA}} & \multicolumn{4}{c}{\textbf{INT}} \\
 & \multicolumn{2}{c}{MWOZ} & \multicolumn{2}{c}{DSTC2} & \multicolumn{2}{c|}{GSIM} & Acc & Acc & Acc & Recall \\
 & $f1_{micro}$ & $f1_{macro}$ & $f1_{micro}$ & $f1_{macro}$ & $f1_{micro}$ & $f1_{macro}$ & (all) & (in) & (out) & (out)  \\
\hline
DSP & \textbf{91.46} & \textbf{80.87} & \textbf{93.26} & \textbf{43.54} & 99.35 & 45.72 & 85.78 & 95.52 & 89.28 & 41.97 \\
ENP & 91.38 & 80.31 & 92.47 & 40.38 & \textbf{99.57} & \textbf{45.85} & \textbf{86.27} & \textbf{95.67} & \textbf{89.60} & \textbf{44.00} \\
\hline
\hline
 & \multicolumn{6}{c|}{\textbf{RS}} & \multicolumn{2}{c}{\textbf{DST}} &  &  \\
 & \multicolumn{2}{c}{MWOZ} & \multicolumn{2}{c}{DSTC2} & \multicolumn{2}{c|}{GSIM} & $acc_{joint}$ & $acc_{slot}$ &  &  \\
 & $R_{100}@1$ & $R_{100}@3$ & $R_{100}@1$ & $R_{100}@3$ & $R_{100}@1$ & $R_{100}@3$ &  &  &  &  \\
\hline
DSP & \textbf{61.79} & \textbf{84.04} & 53.06 & 66.94 & \textbf{36.59} & \textbf{59.37} & 49.02 & 96.96 &  &  \\
ENP & 59.36 & 82.96 & \textbf{54.97} & \textbf{68.83} & 33.33 & 56.17 & \textbf{49.65} & \textbf{97.07} &  & \\
\hline
\end{tabular}
\caption{The experiments investigating the effect of the ability. ENP are designed to learn more abilities which are needed by downstream INT and DST task, and its performance on these two tasks is completely higher than DSP.}
\label{ability_2}
\end{table*}

\begin{table*}[]
\small
\begin{tabular}{l| llllll| llll}
\hline 
 & \multicolumn{6}{c|}{\textbf{DA}} & \multicolumn{4}{c}{\textbf{INT}} \\
 & \multicolumn{2}{c}{MWOZ} & \multicolumn{2}{c}{DSTC2} & \multicolumn{2}{c|}{GSIM} & Acc & Acc & Acc & Recall \\
 & $f1_{micro}$ & $f1_{macro}$ & $f1_{micro}$ & $f1_{macro}$ & $f1_{micro}$ & $f1_{macro}$ & (all) & (in) & (out) & (out)  \\
\hline
DUR & \textbf{91.56} & \textbf{80.16} & \textbf{94.99} & \textbf{45.36} & \textbf{99.48} & \textbf{45.79} & \textbf{85.80} & 95.67 & \textbf{89.11} & \textbf{41.37} \\
CRM & 91.52 & 79.87 & 92.59 & 40.48 & 99.31 & 45.68 & 85.02 & 94.91 & 88.92 & 40.53 \\
\hline
\hline
 & \multicolumn{6}{c|}{\textbf{RS}} & \multicolumn{2}{c}{\textbf{DST}} &  &  \\
 & \multicolumn{2}{c}{MWOZ} & \multicolumn{2}{c}{DSTC2} & \multicolumn{2}{c|}{GSIM} & $acc_{joint}$ & $acc_{slot}$ &  &  \\
 & $R_{100}@1$ & $R_{100}@3$ & $R_{100}@1$ & $R_{100}@3$ & $R_{100}@1$ & $R_{100}@3$ &  &  &  &  \\
\hline
DUR & 59.87 & 83.28 & 55.48 & 69.27 & 29.37 & 53.76 & \textbf{49.36} & \textbf{97.10} &  &  \\
CRM & \textbf{64.27} & \textbf{86.15} & \textbf{58.12} & \textbf{71.86} & \textbf{41.94} & \textbf{66.47} & 48.89 & 96.97 &  & \\
\hline
\end{tabular}
\caption{The experiments investigating the effect of the structure. The model structure of CRM is more similar to downstream task RS, and its performance on this task is completely higher than DUR.}
\label{formulation}
\end{table*}

\subsection{Empirical Analysis on Task Relations between Further Pre-training and Fine-tuning}
In session 3.4, we provide a heuristic analysis on task relations between further pre-training and fine-tuning. We suggest ability and structure are two key factors that influence the effectiveness of further pre-training to fine-tuning.

We define \textbf{nice pair} meaning that a further pre-training task is effective to a downstream task. 
From Table~\ref{question2} we can find DSP is more beneficial for INT, CRM is for RS, while DCV is for DA and DST.
So there are four nice pairs, (DSP, INT), (CRM, RS), (DCV, DA), and (DCV, DST).
Among these four nice pairs, we can find there is one thing in common, \textit{the further pre-training task and downstream task in the same nice pair almost share the same ability and the model structure}. Take (CRM, RS) pair as an example, both CRM and RS mainly learn the ability of dialogue coherence and belong to the siamese structure.
 
To further investigate the effect of the ability, we compare dialogue speaker prediction (DSP) and entity number prediction (ENP). Their structures are the same, that is, single turn classification, but the abilities they learn are different, DSP mainly learns the ability of single turn representation, while ENP also learns entity information. 
Experiment results are shown in Table~\ref{ability_2}, ENP outperforms DRP on intent recognition and dialogue state tracking tasks across all metrics because these two tasks also need the ability about entity information. This indicates ability is important for further pre-training.

To further investigate the effect of the structure, we compare context response matching (CRM) and dialogue utterances reordering (DUR). Both of them mainly learn the ability about dialogue coherence, but their structures are different. Results in Table~\ref{formulation} show that CRM surpasses DUR on the response selection task because the CRM model is a siamese structure which is the same as the response selection task. This indicates the structure is also a crucial factor for the effectiveness of further pre-training.

\subsection{Effect of Combining Further Pre-training Tasks}

\begin{table*}[]
\small
\begin{tabular}{l| llllll| llll}
\hline 
 & \multicolumn{6}{c|}{\textbf{DA}} & \multicolumn{4}{c}{\textbf{INT}} \\
 & \multicolumn{2}{c}{MWOZ} & \multicolumn{2}{c}{DSTC2} & \multicolumn{2}{c|}{GSIM} & Acc & Acc & Acc & Recall \\
 & $f1_{micro}$ & $f1_{macro}$ & $f1_{micro}$ & $f1_{macro}$ & $f1_{micro}$ & $f1_{macro}$ & (all) & (in) & (out) & (out)  \\
\hline
ENP & 91.38 & 80.31 & 92.47 & 40.38 & \textbf{99.57} & 45.85 & \textbf{86.27} & \textbf{95.67} & \textbf{89.60} & \textbf{44.00} \\
CRM & 91.52 & 79.87 & \textbf{92.59} & \textbf{40.48} & 99.31 & 45.68 & 85.02 & 94.91 & 88.92 & 40.53 \\
Joint & \textbf{91.65} & \textbf{80.55} & 92.42 & 40.13 & 99.46 & \textbf{45.78} & 84.31 & 94.96 & 88.26 & 36.40 \\
\hline
\hline
 & \multicolumn{6}{c|}{\textbf{RS}} & \multicolumn{2}{c}{\textbf{DST}} &  &  \\
 & \multicolumn{2}{c}{MWOZ} & \multicolumn{2}{c}{DSTC2} & \multicolumn{2}{c|}{GSIM} & $acc_{joint}$ & $acc_{slot}$ &  &  \\
 & $R_{100}@1$ & $R_{100}@3$ & $R_{100}@1$ & $R_{100}@3$ & $R_{100}@1$ & $R_{100}@3$ &  &  &  &  \\
\hline
ENP & 59.36 & 82.96 & 54.97 & 68.83 & 33.33 & 56.17 & \textbf{49.65} & \textbf{97.07} &  &  \\
CRM & \textbf{64.27} & \textbf{86.15} & \textbf{58.12} & 71.86 & \textbf{41.94} & \textbf{66.47} & 48.89 & 96.97 &  &  \\
Joint & 62.33 & 85.55 & 57.98 & \textbf{72.30} & 39.48 & 64.42 & 49.23 & 97.04 &  & \\
\hline
\end{tabular}
\caption{Results of the experiment investigating the effect of combining multiple further pre-training tasks. The joint model does not improve all the downstream tasks that ENP and CRM beneficial to.} 
\label{joint}
\end{table*}

\begin{table*}[]
\small
\begin{tabular}{l| llllll| llll}
\hline 
 & \multicolumn{6}{c|}{\textbf{DA}} & \multicolumn{4}{c}{\textbf{INT}} \\
 & \multicolumn{2}{c}{MWOZ} & \multicolumn{2}{c}{DSTC2} & \multicolumn{2}{c|}{GSIM} & Acc & Acc & Acc & Recall \\
 & $f1_{micro}$ & $f1_{macro}$ & $f1_{micro}$ & $f1_{macro}$ & $f1_{micro}$ & $f1_{macro}$ & (all) & (in) & (out) & (out)  \\
\hline
ENP & \textbf{91.38} & \textbf{80.31} & \textbf{92.47} & \textbf{40.38} & \textbf{99.57} & \textbf{45.85} & \textbf{86.27} & \textbf{95.67} & \textbf{89.60} & \textbf{44.00} \\
w.o. mlm & 91.06 & 80.27 & 90.75 & 38.16 & 99.34 & 45.72 & 85.79 & 95.57 & 89.18 & 41.77 \\
\hline
\hline
 & \multicolumn{6}{c|}{\textbf{RS}} & \multicolumn{2}{c}{\textbf{DST}} &  &  \\
 & \multicolumn{2}{c}{MWOZ} & \multicolumn{2}{c}{DSTC2} & \multicolumn{2}{c|}{GSIM} & $acc_{joint}$ & $acc_{slot}$ &  &  \\
 & $R_{100}@1$ & $R_{100}@3$ & $R_{100}@1$ & $R_{100}@3$ & $R_{100}@1$ & $R_{100}@3$ &  &  &  &  \\
\hline
ENP & \textbf{59.36} & \textbf{82.96} & \textbf{54.97} & \textbf{68.83} & \textbf{33.33} & \textbf{56.17} & 49.65 & 97.07 &  &  \\
w.o. mlm & 58.09 & 82.29 & 37.12 & 52.98 & 16.92 & 38.70 & \textbf{50.30} & \textbf{97.21} &  &  \\
\hline
\end{tabular}
\caption{The experiment investigating the effect of combining data-level and task-level further pre-training. Removing masked language model objective can cause performance drop on almost all the downstream tasks.}
\label{wo_mlm}
\end{table*}

We jointly further pre-train entity number prediction (ENP) and context response matching (CRM) in the multi-task paradigm, the experiment is denoted as Joint. We expect the joint model can combine the advantages of ENP and CRM, and bring improvement on downstream INT, RS, and DST.
The results in Table~\ref{joint} are not fully consistent with our expectation, specifically, on intent recognition, joint model's performance drops significantly, on the other three downstream tasks, joint model's performance is between ENP and CRM. 

\subsection{Effect of Combining Data-level and Task-level Further Pre-training}
In the former experiments, each proposed further pre-trained task is trained with masked language model (MLM), we suppose MLM is for data-level adaptation while the proposed task is for task-level adaptation.
In this section, we investigate the effect of MLM by removing MLM objective from further pre-training stage, this experiment is denoted as w.o. mlm.
Experiment results are shown in Table~\ref{wo_mlm}. Removing MLM leads to performance drop across almost all downstream tasks, indicating MLM is important to further pre-training stage.

\subsection{Experiment Summary}

Through all the experiments, we can conclude that: In the area of task-oriented dialogue, 
1) Masked language model alone is not enough for further pre-training, but it still plays an important role for enhancing fine-tuning. And there is a need to design special further pre-training tasks according to the characteristics of dialogue data.
2) Different pre-training tasks do have different effects on different downstream tasks, and it is necessary to design a specific pre-training task for a specific downstream task.
3) Ability and structure of a further pre-training task are key factors influencing the performance of fine-tuning on a downstream task.
4) Training two further pre-training tasks in the multi-task paradigm does not lead to incremental performance improvement.

From these conclusions, we can obtain multiple empirical principles to design further pre-training tasks: 1) The ability learned by pre-training task should be similar to the ability required by the downstream task. 2) the modeling structure should also be similar, 3) the masked language model training objective should be kept. 

\section{Conclusion}

In this work, we study how to make further pre-training more effective to downstream tasks in the area of the task-oriented dialog.
Firstly, we notice that further pre-training using MLM objective does not improve all downstream tasks, 
then we designed multiple pre-training tasks for dialog data, finding that different pre-training tasks benefit different downstream tasks. 
Further, we observe that ability and structure are key factors influencing whether a pre-training task is helpful to a downstream task. 
These finds can be used as empirical principles to design pre-training tasks.

\section*{Acknowledgments}
We would like to thank all the reviewers for their insightful and valuable comments and suggestions.


\bibliography{custom}
\bibliographystyle{acl_natbib}

\end{document}